\documentclass[10pt,twocolumn,letterpaper]{article}

\usepackage[pagenumbers]{cvpr} 

\usepackage{multirow}

\usepackage{placeins}

\definecolor{cvprblue}{rgb}{0.21,0.49,0.74}
\usepackage[pagebackref,breaklinks,colorlinks,allcolors=cvprblue]{hyperref}
\usepackage[accsupp]{axessibility}  

\usepackage[normalem]{ulem}

\usepackage{amssymb}
\usepackage{booktabs}
\usepackage{pifont}

\usepackage{colortbl}
\usepackage{multirow}
\usepackage{float}

\newcommand{\calD}{{\mathcal{D}}}

\newcommand{\calL}{{\mathcal{L}}}

\newcommand{\be}{\begin{eqnarray}}
\newcommand{\ee}{\end{eqnarray}}
\newcommand{\bee}{\begin{eqnarray*}}
\newcommand{\eee}{\end{eqnarray*}}

\newcommand{\matrixb}{\left[ \begin{array}}
\newcommand{\matrixe}{\end{array} \right]}

\newcommand{\app}{\raise.17ex\hbox{$\scriptstyle\sim$}}
\input{def_custom.tex}

\newcommand{\newpara}[1]{\vspace{6pt}\noindent\textbf{#1}}

\newcommand\nonumberfootnote[1]{%
  \begingroup%
  \let\thefootnote\relax%
  \footnotetext{#1}%
  \addtocounter{footnote}{-1}%
  \endgroup%
}

\def\paperID{15779} 
\def\confName{CVPR}
\def\confYear{2025}

\title{Seeing Speech and Sound: Distinguishing and Locating Audios in Visual Scenes
}

\author{Hyeonggon Ryu{*} \quad Seongyu Kim{*} \quad Joon Son Chung \quad Arda Senocak\\
Korea Advanced Institute of Science and Technology\\
}

\begin{document}
\twocolumn[{%
\renewcommand\twocolumn[1][]{#1}%
\maketitle
\setcounter{footnote}{1}
\begin{center}
    \centering
    \vspace{-.2in}
    \captionsetup{type=figure}
    \includegraphics[width=\textwidth]{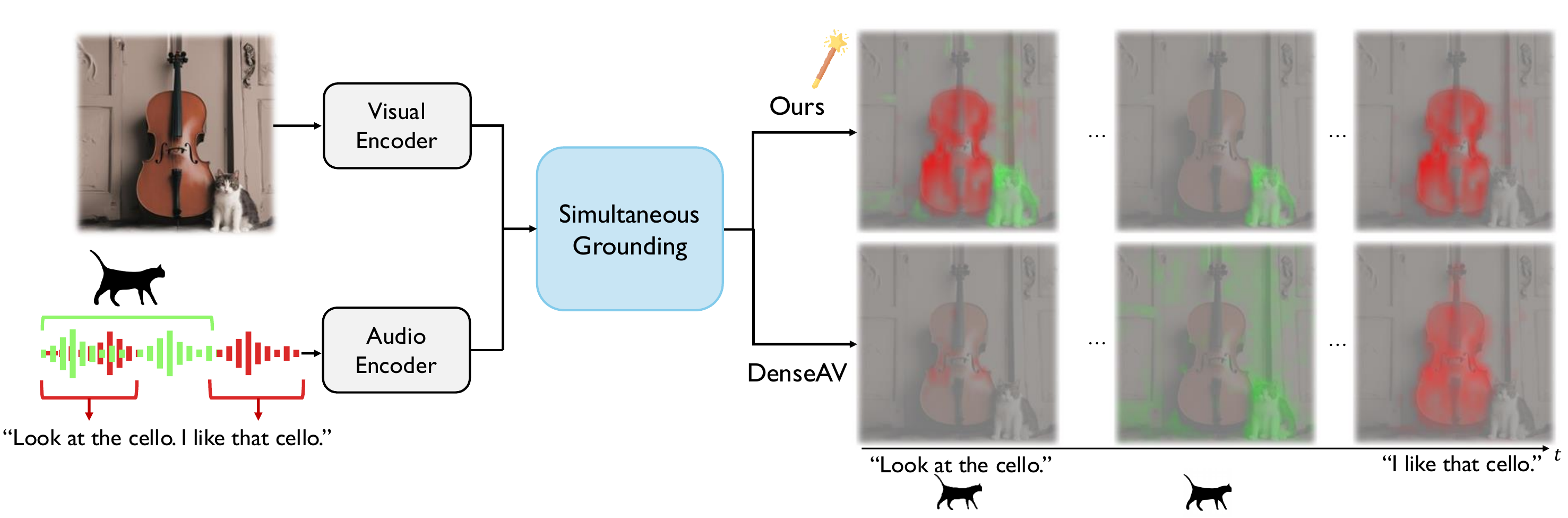}
    \vspace{-.3in}
    \caption{\textbf{Simultaneous Audio-Visual Grounding.} 
    We introduce a model for simultaneous grounding of mixed audio types — including \textit{overlapping} spoken language and non-speech sounds — within a visual scene. Our framework takes audio-visual grounding to the next level, overcoming the limitations of existing methods that handle audio types independently or sequentially without overlapping.
    }
    \label{fig:teaser}
\end{center}%
}]
\nonumberfootnote{$^{*}$These authors contributed equally to this work.}
\begin{abstract}

We present a unified model capable of simultaneously grounding both spoken language and non-speech sounds within a visual scene, addressing key limitations in current audio-visual grounding models. 
Existing approaches are typically limited to handling either speech or non-speech sounds independently, or at best, together but sequentially without mixing. This limitation prevents them from capturing the complexity of real-world audio sources that are often mixed. Our approach introduces a ``mix-and-separate'' framework with audio-visual alignment objectives that jointly learn correspondence and disentanglement using mixed audio.  
Through these objectives, our model learns to produce distinct embeddings for each audio type, enabling effective disentanglement and grounding across mixed audio sources.
Additionally, we created a new dataset to evaluate simultaneous grounding of mixed audio sources, demonstrating that our model outperforms prior methods. Our approach also achieves comparable or better performance in standard segmentation and cross-modal retrieval tasks, highlighting the benefits of our mix-and-separate approach.
\end{abstract}    
\section{Introduction}
\label{sec:intro}

We live in a world rich with multi-sensory signals. Our interactions with it rely on our natural multimodal perception, drawing on auditory and visual cues as well as spoken language to gain a complete understanding of our surroundings. While we can actively limit or focus our field of view in visual scenes, auditory signals are a mixture of different types of sounds, such as speech, music, and a variety of environmental or object sounds, all entering our ears as a combined mix without any filtering. Despite this complexity, humans can still discern specific types of sounds and associate them with their visual counterparts — for example, the sound of a ``lion roaring'', someone saying the word ``lion'', or the song ``Nants' Ingonyama''\footnote{
Nants' Ingonyama is the iconic opening song from The Lion King
\url{https://www.youtube.com/watch?v=CF-c1K3WWg4}} linked with the visual of a \textit{lion}. Remarkably, we do this without any direct supervision.

This natural audio-visual association enables us to locate which object a spoken word refers to or identify which object is producing a sound within a visual scene.  
Inspired by the human ability to learn these localization skills from natural audio-visual correspondences without explicit supervision, extensive studies~\cite{senocak2018learning,senocak2019learning,senocak2022learning,senocak2023sound,senocak2024aligning,mo2022localizing,mo2022SLAVC,harwath2018vision,PengFastVGS22,PengWordDiscovery22} have aimed to replicate this understanding in machine perception. However, most existing methods focus on understanding relationships between images and either spoken utterances or generic sounds, where each approach handles only one type of audio at a time but not both together.

Recently, Hamilton \etal~\cite{hamilton2024separating} proposed a model that visually grounds both spoken language and non-speech sounds within an image scene. This model is trained to disentangle different types of audio, such as environmental sounds or spoken words, and to locate their sources within a visual context, all without supervision. While pioneering, this work has several limitations. 
First, the model handles either speech or non-speech audio individually, rather than mixed audio sources, limiting its ability to process real-world audio, which often contains overlapping sounds. 
Second, the model does not achieve effective disentanglement in distinguishing between different types of auditory signals (\eg, speech and generic sounds). Although each of the heads in the audio encoder primarily activates for a specific audio type, these activations function as a mere selection mechanism for each type, rather than effectively controlling the flow of information between types — resulting in information leakage. 
Consequently, the multi-heads learn features of the opposite audio type, which impairs disentanglement and makes the model vulnerable when confronted with mixed audio inputs.

To address these limitations, we propose a unified method that \textit{simultaneously} grounds both spoken language and non-speech sounds within a visual scene (~\Fref{fig:teaser}). Our model tackles both challenges using a ``mix-and-separate'' approach. We introduce a novel mixed audio-visual alignment objective that jointly learns \textit{correspondence} and \textit{disentanglement}, aligning representations of multiple audio types with corresponding visuals in images. Our framework includes shared audio and visual encoders within each modality, facilitating the transfer of audio-visual interactions across learning objectives. Unlike traditional mix-and-separate approaches, we do not generate separated audio; instead, we obtain embeddings from each specialized head of the audio encoder that represent the disentangled sound sources of different types using contrastive learning. These learning objectives enable our framework to process audio mixtures, eliminating the need for multiple audio types to appear sequentially or individually during training and inference. Additionally, this approach improves disentanglement within the multi-head structure.

The primary motivation of this paper is to achieve simultaneous grounding of multiple audio types; accordingly, we aim to evaluate our model’s ability to perform this task. However, no existing dataset is suitable for this purpose. To address this, we created a new dataset by extending the IS3 dataset~\cite{senocak2024aligning} — which provides segmentation masks for multiple sound-producing objects with corresponding audio — by adding spoken utterances to each sample. Evaluation on this dataset demonstrates that our model outperforms existing methods in simultaneous sound and speech grounding, a task that requires handling mixed audio inputs. Additionally, by achieving superior performance on standard segmentation and cross-modal retrieval experiments on the datasets from~\cite{hamilton2024separating}, we show the benefits of our self-supervised mix-and-separate approach.

Our main contributions are as follows:
\begin{itemize}
    \item We propose a method for \textit{simultaneously} grounding both speech and non-speech audios within a visual scene.
    \item We introduce a novel 
    audio-visual alignment objective that uses \textit{audio mixtures} and jointly learns \textit{correspondence} and \textit{disentanglement} using contrastive learning.
    \item We provide a new dataset, \textit{Extended-IS3},  for simultaneous sound and speech grounding and demonstrate that our model outperforms existing methods.
\end{itemize}

\section{Related Works}
\label{sec:related_works}
\newpara{Sound source localization.} Audio-visual sound source localization is the task of identifying objects, events, or areas within a visual scene based on corresponding audio cues. In the deep learning era, Senocak \etal~\cite{senocak2018learning,senocak2019learning} pioneered this field by introducing a cross-modal attention mechanism with contrastive learning to align audio and visual information. As this research field advanced, diverse techniques have emerged, including hard sample mining~\cite{chen2021localizing}, 
multiple-instance contrastive learning~\cite{mo2022localizing}, negative-free strategies~\cite{song2022sspl}, momentum encoders~\cite{mo2022SLAVC}, intra-modality similarity learning~\cite{sun2023learning}, and the use of multiple positives in contrastive learning~\cite{senocak2023sound}. Recently, CLIP has also been incorporated into sound source localization without using text~\cite{park2024can}. Most prior methods focus on single-source localization, while another line of research addresses multiple sound source localization with generic sounds~\cite{hu2020discriminative,qian2020multiple,hu2022mix,mo2023audiovisual,mahmud2024t,kim2024learning}.  
In contrast to these models, which primarily localize generic sounds, our work targets both generic sounds and spoken utterances, either simultaneously or individually, with a single model. Our approach addresses the added complexity of mixed generic sounds and spoken language, requiring a deeper understanding of language and its integration with visual scenes.

\begin{figure*}[t!]
    \centering
    \includegraphics[width=0.95\linewidth]{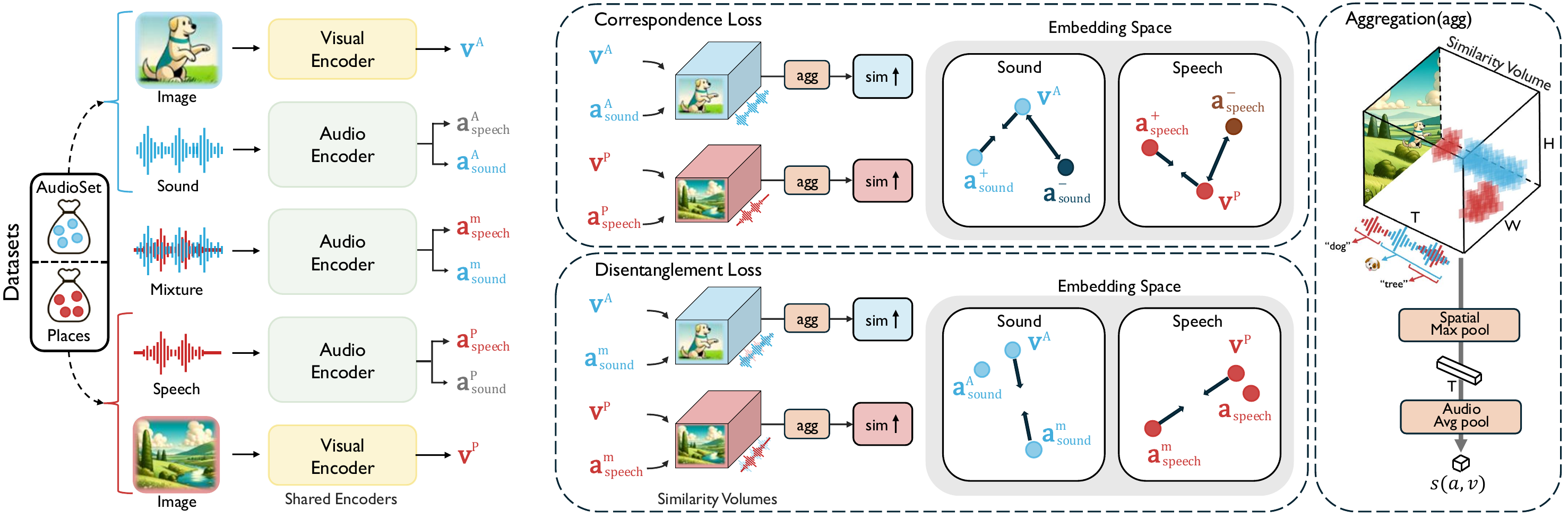}
    \vspace{-2mm}
    \caption{\textbf{The pipeline of our framework.} 
    Visual and audio encoders extract features from clean audios, their paired images, and mixed audio inputs. These features are used to compute audio-visual similarities, which are then used in our correspondence and disentanglement losses. The correspondence objective ensures audio-visual matching, while the disentanglement objective enables feature-level separation of mixed audio sources.}
    \label{fig:framework_figure}
    \vspace{-4mm}
\end{figure*}

\newpara{Visually grounded speech.} Visually grounded speech (VGS) aims to associate images with corresponding spoken utterances. Various tasks have been explored in this domain, including cross-modal retrieval~\cite{PengFastVGS22,woo2024speech}, multilingual associations across languages using images as an interlingua~\cite{harwath2018vision, Ohishi2020Pair, ohishi2020trilingual,ryu2023hindi}, speech unit or word discovery and keyword spotting~\cite{PengWordDiscovery22, Harwath2020LearningHD, shih2023speechclip}, and visual grounding~\cite{harwath2018jointly}. Our approach aligns with the visual grounding work of~\cite{harwath2018jointly}; however, it differentiates itself from previous work by enabling the grounding of both spoken utterances and generic sounds within a single model. This broader capability allows our model to handle diverse audio inputs, advancing VGS beyond isolated spoken language associations.

\newpara{Audio-visual sound source separation.} Audio-only sound source separation has a long history, and with the advances in deep learning, audio-visual sound source separation has become an established field. Audio-visual separation methods leverage visual cues to isolate individual sound sources, often employing a mix-and-separate approach. As a pioneering work, Zhao \etal~\cite{zhao2018sound} introduced a method that uses pixel-level correspondence between visual information and audio to separate sound sources in images. Subsequent approaches have incorporated additional cues, such as scene graphs~\cite{chatterjee2021visual}, motion~\cite{zhao2019sound}, gestures and keypoints~\cite{gan2020music}, to enhance separation performance. The majority of these works focus on general sounds, such as objects or musical instruments, while other lines of research address multi-speaker speech separation~\cite{owens2018audio, gao2021visualvoice} and distinguishing on-screen sounds from off-screen sounds~\cite{owens2018audio, tzinis2020into}. While most prior work focuses exclusively on audio separation,~\cite{tian2021cyclic,mo2023unified} propose models that jointly optimize separation and localization within a unified architecture for generic sounds only. Like previous approaches, our method employs a mix-and-separate strategy; however, rather than generating separated audio, we perform separation at the feature level, yielding features that represent distinct sound sources. Additionally, whereas existing models typically specialize in either speech or specific type of general sounds, our approach introduces a unified model capable of separating and localizing both generic sounds and spoken language, either simultaneously or individually, within a single framework.
\section{Methodology}
\label{sec:formatting}
Our goal is to perform \textit{simultaneous} grounding of both speech and non-speech audio within a visual scene using a single model. To this end, we propose a self-supervised “mix-and-separate” strategy. Given an audio mixture and its corresponding image, the model parses the audio into speech and sound sources, and then localizes them within the image. We frame this task as a representation learning problem, using two contrastive losses—correspondence and disentanglement—which are jointly optimized to transfer complementary information across objectives. Our proposed pipeline is shown in~\Fref{fig:framework_figure}.

\subsection{Preliminaries}
\newpara{Contrastive learning} aims to learn representations by distinguishing between positive and negative pairs. In audio-visual learning, for an encoded audio sample $\mathbf{a}_i=f_a(a_i)$ and its corresponding positive image feature pair $\mathbf{v}_i=f_v(v_i)$, with negative pairs $\mathbf{v}_j$ for $i \neq j$, from a dataset $\calD=\{(v_i,a_i):i=1,...,N\}$, where $v_i$ and $a_i$ are image and audio sample, $f_v$ and $f_a$ are image and audio encoder respectively, the loss can be formulated as:
\begin{equation}
\label{eq:av_contrastive}
\begin{aligned}
        \calL_i=\mathrm{-log}\frac{\mathrm{exp}(s(\mathbf{a}_i, \mathbf{v}_i)/\tau)}{\sum_{j}\mathrm{exp}(s(\mathbf{a}_i, \mathbf{v}_j)/\tau)},
\end{aligned}
\end{equation}
where $s$ refers to the cross modal similarity function and $\tau$ is a trainable temperature parameter. The contrastive loss~\Eref{eq:av_contrastive} is applied symmetrically by following previous works~\cite{hamilton2024separating,mo2022localizing,sung2023sound}.

\newpara{Audio and visual encoders with aligner.} Given an image $v_i$ with its corresponding audio $a_i$, our backbone networks extract features for each modality. The vision backbone $f_v$ processes an image frame to extract a spatial feature map, $\mathbf{v_d} \in \mathbb{R}^{C_D \times H \times W}$, while the audio backbone $f_a$ processes the raw waveform to extract an audio feature, $\mathbf{a_h} \in \mathbb{R}^{C_H \times F \times T}$. Each modality encoder has its own aligner network attached at the end of the backbones. These aligners consist of a Channel-wise LayerNorm block and convolution layers, which reshapes the image spatial feature map to 
$\mathbf{v} \in \mathbb{R}^{C \times K \times H \times W}$ and the audio feature to $\mathbf{a} \in \mathbb{R}^{C \times K \times F \times T}$. Here, \( K \) denotes the number of audio types (set to 2), meaning there is a specific head for each audio type, while \( F \), \( T \), \( H\) and \( W\) represent the frequency, temporal, and spatial axes, respectively. $C_D$, $C_H$ and $C$ refers to channel dimension of vision encoder, audio encoder and shared feature space respectively.

\newpara{Similarity function.} To obtain distinct features representing each audio sub-modality, it is essential to use a function that computes the similarity between audio and image with consideration of the audio type -- the two embedding spaces for speech and sound. 
We then build a similarity volume $S\mathbf{(a, v)}\in \mathbb{R}^{K \times F \times T \times H \times W}$ from the image spatial feature map and the audio feature:
\vspace{-2mm}
\begin{align}
S(\mathbf{a}, \mathbf{v}) &= \sum_{c=1}^{C} \mathbf{a}[c, k, f, t] \cdot \mathbf{v}[c, k, h, w],\label{eq:similarity_volume}
\end{align} where $\mathbf{a}[c, k, f, t]$ refers to the value of $\mathbf{a}$ at location $[c, k, f, t]$ and · denotes scalar multiplication. Thus, each audio type has its own similarity volume with the corresponding image. Aggregation across the head axis can be done in two different ways, either through max pooling or head selection:
\begin{align}
S_M(\mathbf{a}, \mathbf{v}) &= \max_{k} \left( S(\mathbf{a}, \mathbf{v}) \right),
\label{eq:selection_methods_max}\\
S_N(\mathbf{a}, \mathbf{v}) &= S(\mathbf{a}, \mathbf{v})[N],
\label{eq:selection_methods_indexing}
\end{align}
where $[N]$ follows PyTorch slicing to select activations for the \textit{N-th} audio type head. After aggregation along the head axis, spatial max pooling is applied, followed by average pooling across the frequency and temporal axes to define the similarity score $s_M(\mathbf{a}, \mathbf{v})$ or $s_N(\mathbf{a}, \mathbf{v})$ (shown in~\Fref{fig:framework_figure}):
\begin{align}
s_M(\mathbf{a}, \mathbf{v}) &= \frac{1}{FT} \sum_{f=1}^{F} \sum_{t=1}^{T} \max_{h, w} \left( S_M(\mathbf{a}, \mathbf{v})[f, t, h, w] \right),\label{similarity_score_max}
\end{align}
\vspace{-2mm}
\begin{align}
s_N(\mathbf{a}, \mathbf{v}) &= \frac{1}{FT} \sum_{f=1}^{F} \sum_{t=1}^{T} \max_{h, w} \left( S_N(\mathbf{a}, \mathbf{v})[f, t, h, w] \right),\label{similarity_score_indexing}
\end{align}
where we use either $N$ or $N'$ for $s_N(\mathbf{a}, \mathbf{v})$, denoting the index of the head specialized in sound and speech respectively.

\subsection{Mix-and-Separate}
Our method employs a ``mix-and-separate'' approach with a novel mixed audio-visual alignment objectives that jointly learn both \textit{correspondence} and \textit{disentanglement}, aligning representations of multiple audio types with their corresponding visuals in images.

\newpara{Correspondence loss} is designed to establish shared audio-visual feature spaces. By using clean sound or speech samples as input audio, the correspondence loss encourages the model to build associations between each audio sample and its corresponding image pair. Given the clean audio feature of sound and speech, and the paired image feature map of each audio type, denoted as $\mathbf{a}^\texttt{{A}}$, $\mathbf{a}^{\texttt{P}}$, $\mathbf{v}^\texttt{A}$, and $\mathbf{v}^\texttt{P}$, respectively, the correspondence loss is formulated as follows:
\begin{equation}
\label{eq:l_correspondence}
\begin{aligned}
        \calL_{cor}={\sum_{i}\mathrm{-log}\frac{\mathrm{exp}(s_M(\mathbf{a}^{\texttt{A}}_i, \mathbf{v}^\texttt{A}_i)/\tau)}{\sum_{j}\mathrm{exp}(s_M(\mathbf{a}^{\texttt{A}}_i, \mathbf{v}^\texttt{A}_j)/\tau)}}\\
        +{\sum_{i}\mathrm{-log}\frac{\mathrm{exp}(s_M(\mathbf{a}^{\texttt{P}}_i, \mathbf{v}^\texttt{P}_i)/\tau)}{\sum_{j}\mathrm{exp}(s_M(\mathbf{a}^{\texttt{P}}_i, \mathbf{v}^\texttt{P}_j)/\tau)}}.
\end{aligned}
\end{equation}

We visualize the features in~\Fref{fig:framework_figure} for better understanding:\\
\vspace{-3mm}
\begin{equation}
\label{eq:visualize_clean_feature}
\begin{aligned}
\mathbf{a}^\texttt{{A}}_\texttt{sound}&=\mathbf{a}^\texttt{{A}}[N], 
\mathbf{a}^\texttt{{A}}_\texttt{speech}=
\mathbf{a}^\texttt{{A}}[N'],\\
\mathbf{a}^\texttt{{P}}_\texttt{sound}&=\mathbf{a}^\texttt{{P}}[N],  
\mathbf{a}^\texttt{{P}}_\texttt{speech}=\mathbf{a}^\texttt{{P}}[N'].
\end{aligned}
\end{equation}
\\
\newpara{Disentanglement loss} in our model is designed to ensure the disentanglement of audio type heads. To achieve this, we adopt a \textit{mix-and-separate} strategy that uses audio mixtures containing both audio types. The audio encoder generates distinct features for each sub-modality, represented by two separate audio heads that align with features derived from paired image samples for both speech and sound. In this audio-visual feature space, each image feature is already close to its corresponding clean audio feature. The model learns disentanglement by aligning the features from the mixed audio with the corresponding image features and, inherently, with the features from the clean audios as shown in~\Fref{fig:framework_figure}. 
Given the feature of mixed audio that is a combination of sound and speech, $\mathbf{a}_\texttt{{m}}$, and the paired image feature map from each of sound and speech, $\mathbf{v}_\texttt{A}$, and $\mathbf{v}_\texttt{P}$, the disentanglement loss is formulated as follows:
 
\begin{equation}
\label{eq:l_disentangle}
\begin{aligned}
        \calL_{dis}={\sum_{i}\mathrm{-log}\frac{\mathrm{exp}(s_N(\mathbf{a}^{\texttt{m}}_i, \mathbf{v}^\texttt{A}_i)/\tau)}{\sum_{j}\mathrm{exp}(s_N(\mathbf{a}^{\texttt{m}}_i, \mathbf{v}^\texttt{A}_j)/\tau)}}\\
        +{\sum_{i}\mathrm{-log}\frac{\mathrm{exp}(s_{N'}(\mathbf{a}^{\texttt{m}}_i, \mathbf{v}^\texttt{P}_i)/\tau)}{\sum_{j}\mathrm{exp}(s_{N'}(\mathbf{a}^{\texttt{m}}_i, \mathbf{v}^\texttt{P}_j)/\tau)}}.
\end{aligned}
\end{equation}

The features are visualized in~\Fref{fig:framework_figure} for better understanding:\\
\vspace{-3mm}
\begin{equation}
\label{eq:visualize_mixed_feature}
\begin{aligned}
\mathbf{a}^\texttt{{m}}_\texttt{sound}&=\mathbf{a}^\texttt{{m}}[N],  
\mathbf{a}^\texttt{{m}}_\texttt{speech}=
\mathbf{a}^\texttt{{m}}[N'].\\
\end{aligned}
\end{equation}

\newpara{Total loss} is formulated as a mixed audio-visual alignment objective to jointly achieve both correspondence and disentanglement, as follows:
\begin{equation}
\label{eq:l_total}
\begin{aligned}
        \calL_{total}=\calL_{cor}+\calL_{dis}.
\end{aligned}
\end{equation}

Our proposed model is trained end-to-end together with this training objective.

\subsection{Implementation and Training Details}

\subsubsection{Architecture Details}
\newpara{Image encoder $f_v(\cdot)$.} We use DINO~\cite{caron2021emerging}, pre-trained on ImageNet in a self-supervised manner. An aligner module, consisting of a Channel-wise LayerNorm block and convolutional layers, is added to the end of DINO. We apply LoRA~\cite{hu2021lora} to fine-tune only the attention blocks of DINO in our audio-visual end-to-end training objective.

\newpara{Audio encoder $f_a(\cdot)$.} We use HuBERT Large~\cite{hsu2021hubert}, pre-trained on Libri-Light in a self-supervised manner. HuBERT is one of the de facto models in speech processing, specifically designed and pre-trained for speech. An aligner module, consisting of a Channel-wise LayerNorm block and convolutional layers, is added to the end of HuBERT. The audio encoder is fully fine-tuned within our audio-visual end-to-end training objective.

\subsubsection{Implementation Details}
Inputs to our model consist of a single 224 $\texttt{x}$ 224 image and a 10-second audio waveform sampled at 16 kHz. We preprocess images and audio following prior work~\cite{hamilton2024separating}, applying standard image augmentations and trimming or padding audio to 10 seconds. The model is trained for 800K steps on 8 A6000 GPUs with an effective batch size of 64, using a LoRA rank of 8 for image encoder. Additional details are provided in the supplementary material.

\newpara{Other details.} To ensure stable model convergence, we incorporate two additional techniques, following~\cite{hamilton2024separating}. First, a warm-up stage for the aligners — attached to each modality encoder — is applied, during which the image and audio backbone networks remain frozen, and only the aligners are trained for 3K steps using our correspondence loss objective. Once the warm-up stage ends, the entire model is trained end-to-end as described before. Second, additional regularizer losses are applied to enhance training stability. Please refer to the supp. material and~\cite{hamilton2024separating} for further details. We emphasize that while these techniques do not significantly impact the model's performance, they contribute to more stable training.
\section{Experiments}

\subsection{Datasets}\label{ssec:datasets}

\newpara{Training datasets.} Our model is trained using the AudioSet~\cite{gemmeke2017audio} and PlacesAudio~\cite{harwath2016unsupervised} datasets. AudioSet is a large-scale audio-visual dataset containing \app 2M samples of general sounds, while PlacesAudio includes \app 400K samples of spoken utterances paired with images. Since our method processes mixture of audios, we combine each PlacesAudio sample with a randomly selected AudioSet sample, using their respective images as part of the training process (see~\Fref{fig:framework_figure}).

\newpara{Testing datasets.} After training, we test the grounding and retrieval performances with the datasets below.
\begin{itemize}
    \item \textbf{Test sets from DenseAV~\cite{hamilton2024separating}.} For standard audio-visual grounding, we use the ADE20K Speech and Sound dataset from~\cite{hamilton2024separating}, which includes 3,030 image-audio pairs for speech and 106 for sound. For retrieval tasks, we use the PlacesAudio test set and a 1,000-sample subset of the AudioSet validation set, selected randomly following~\cite{hamilton2024separating}.
    
    \item \textbf{Extended IS3~\cite{senocak2024aligning}.} The IS3 Dataset is a test set consisting of 3,420 images for evaluating interactive sound source localization, where two objects are present in the visual scene. Each image is paired with two audio samples that represent each visible object. We extend this dataset by generating two individual speech audio samples for each image, each reading the class category of the sounding objects in the scene. This extension provides every image with both the generic sounds of visible objects and their spoken labels. The speech samples are generated using the Google gTTS API. This new dataset enables us to assess whether models can simultaneously ground both audio types within a mixture.

\end{itemize}

\subsection{Baselines}\label{ssec:baselines}
We compare our method against the following baselines. ImageBind~\cite{girdhar2023imagebind} is a multimodal foundation model trained on large-scale datasets across several modalities, including text. CAV-MAE~\cite{gong2023contrastive} is a state-of-the-art model designed for audio-visual learning, achieving strong performance on tasks like audio-visual retrieval. 
DenseAV~\cite{hamilton2024separating} is the most recent and closely related work, as discussed earlier. We report results using the released checkpoints of this model from its official page.

\subsection{Main Results} \label{ssec:quan_qual}
In this section, we present results comparing our proposed method with prior and closely related works on simultaneous audio-visual grounding, speech or generic sound prompted visual grounding, and cross-modal retrieval tasks.

\begin{figure*}[t!]
    \centering
    \includegraphics[width=1\linewidth]{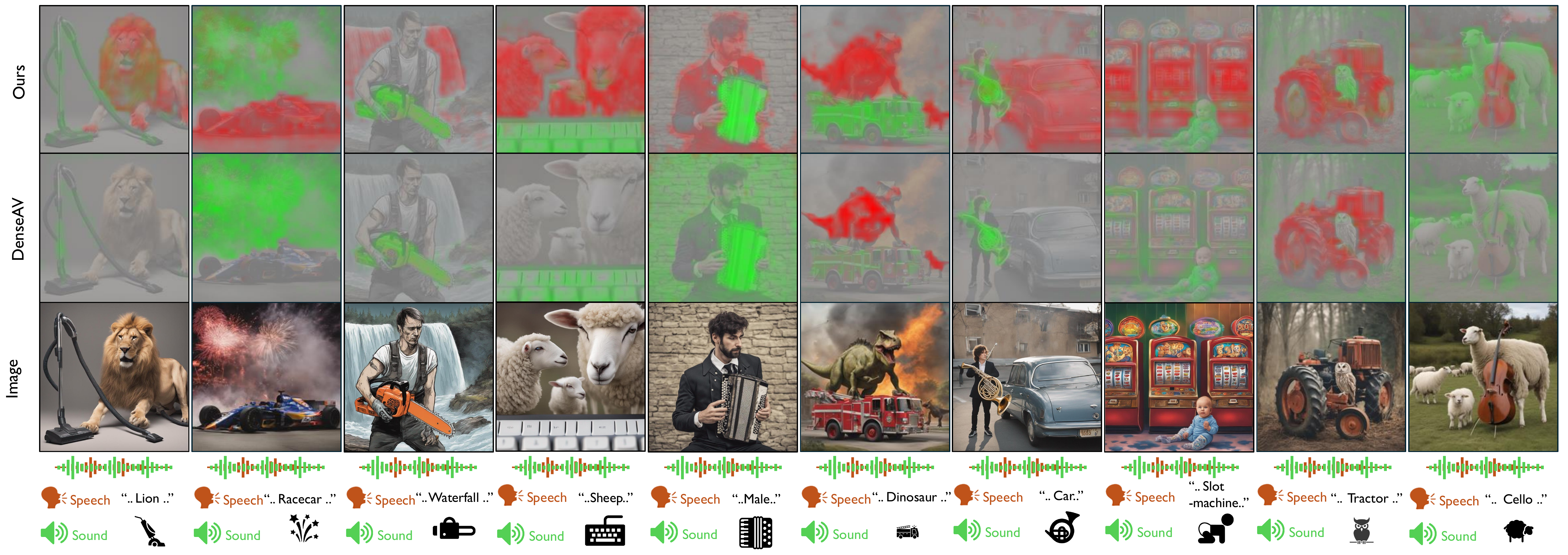}
    \vspace{-6mm}
    \caption{\textbf{Qualitative results for simultaneous audio-visual grounding on Extended IS3 dataset.} Our model accurately localizes both overlapping audio types simultaneously within the mixed audio, whereas competing method~\cite{hamilton2024separating} cannot.}
    \label{fig:is3_simultaneous}
    \vspace{-4mm}
\end{figure*}

\subsubsection{Simultaneous Audio-Visual Grounding} \label{ssec:simultaneous_grounding}
The main goal of this paper is
\textit{simultaneously} grounding both speech and non-speech audios within a visual scene; therefore, in this section, we aim to evaluate our model's ability to perform this task.

\newpara{Implementation.} In this experiment, we use our \textit{Extended IS3} dataset. Audio input is a mixed audio that is the summation of speech and sound. The evaluation metrics are mean average precision (mAP) and intersection over union (mIoU). Simultaneous semantic segmentation results for each audio type are obtained using the corresponding head.

\newpara{Quantitative results.} We present our results in~\Tref{tab:is3_semseg_results}, demonstrating that our model outperforms the recent work~\cite{hamilton2024separating}, DenseAV, by \textbf{+12.7 mAP} and \textbf{+5.2 mAP} for Sound and Speech segmentation, respectively, in the simultaneous grounding scenario. 
These results indicate that our proposed model successfully learns to localize distinct audio types simultaneously through our mix-and-separate approach and specific learning objectives.

\begin{table}[t]
\centering
\resizebox{1.0\linewidth}{!}{  
\begin{tabular}{lcccc}
\toprule
\textbf{Method}         & \multicolumn{2}{c}{\textbf{Sim-Sound Seg.}} & \multicolumn{2}{c}{\textbf{Sim-Speech Seg.}} \\ \cmidrule(lr){2-3} \cmidrule(lr){4-5}
& \textbf{mAP}  & \textbf{mIoU}  & \textbf{mAP}  & \textbf{mIoU}     
\\ \midrule
\textcolor{lightgray}{\textit{Foundation Model}} &  &  &  &    \\
\textcolor{lightgray}{ImageBind~\cite{girdhar2023imagebind}$_{\text{CVPR}23}$} & \textcolor{lightgray}{17.9} & \textcolor{lightgray}{19.2} & \textcolor{lightgray}{24.2} & \textcolor{lightgray}{22.6} \\
\midrule
CAVMAE~\cite{gong2023contrastive}$_{\text{ICLR}23}$  & 31.0 & 27.9  & 15.6 & 19.0 \\
DenseAV~\cite{hamilton2024separating}$_{\text{CVPR}24}$  & 30.7 & 27.0 & 53.4 & 44.0 \\
\textbf{Ours}  & \textbf{43.4} & \textbf{34.6} & \textbf{58.6} & \textbf{47.0} \\ 
\bottomrule
\end{tabular}
}
\caption{\textbf{Simultaneous Audio-Visual segmentation results.}}
\vspace{-4mm}
\label{tab:is3_semseg_results}
\end{table}

\newpara{Qualitative results.} To further evaluate our method, we visualize and compare our simultaneous audio-visual segmentation results with recent prior work. The visual samples in~\Fref{fig:is3_simultaneous} again demonstrate that our model accurately grounds both spoken words and sound-producing objects, even when speech and generic sounds overlap.

\subsubsection{Individual Audio-Visual Grounding} \label{ssec:grounding}
Here, we validate the effectiveness of our proposed method in the standard semantic segmentation task as in~\cite{hamilton2024separating}, using two types of audio individually: speech and generic sound. The task here is that, for a given audio input, the model generates attention maps for the \textit{meaning} of the spoken word or the \textit{object} that produces the sound.

\newpara{Implementation.} 
We follow the evaluation protocol in DenseAV~\cite{hamilton2024separating}, using the same test sets as in~\cite{hamilton2024separating} (~\Sref{ssec:datasets}) and employing mAP and mIoU as evaluation metrics. Our model is evaluated with two aggregation approaches across the heads: \textit{total} and \textit{specialized head}. The \textit{total} approach uses element-wise max pooling across all heads, while the \textit{specialized head} approach selects the head specialized for the target audio type. For speech grounding, we average activations over the duration of the ``[object]'' in the prompt ``A picture of a(n) [object]'', using word timing information from the speech.

\newpara{Results on  sound prompted visual grounding.} In this experiment, we conduct a comparative analysis of our proposed method using only generic sound. We present our results in~\Tref{tab:clean_semseg_results}. Our method achieves improvements of \textbf{+6.4 mAP} and \textbf{+1.9 mAP} in the \textit{Total} and \textit{Sound Head} evaluations, respectively, compared to previous work~\cite{hamilton2024separating}. This is particularly important given that the audio backbone network is HuBERT~\cite{hsu2021hubert}, one of the de facto models in speech processing, specifically designed and pre-trained for speech. Adapting HuBERT to better understand generic sounds demonstrates the effectiveness of our approach.
\begin{table}[t]
\centering
\resizebox{1.0\linewidth}{!}{  
\begin{tabular}{lcccccccc}
\toprule
\textbf{Method} & \multicolumn{8}{c}{\textbf{Clean SemSeg}} \\ \cmidrule(lr){2-9}
& \multicolumn{4}{c}{\textbf{Sound SemSeg}} & \multicolumn{4}{c}{\textbf{Speech SemSeg}} \\ \cmidrule(lr){2-5} \cmidrule(lr){6-9}
&\multicolumn{2}{c}{\textbf{Total}} &\multicolumn{2}{c}{\textbf{Sound}}
&\multicolumn{2}{c}{\textbf{Total}} &\multicolumn{2}{c}{\textbf{Speech}} \\
& \textbf{mAP} & \textbf{mIoU} & \textbf{mAP} & \textbf{mIoU} & \textbf{mAP} & \textbf{mIoU} & \textbf{mAP} & \textbf{mIoU} \\ 
\midrule      
\textcolor{lightgray}{\textit{Foundation Model}} &  &  &  &  &  &  &  &    \\
\textcolor{lightgray}{ImageBind~\cite{girdhar2023imagebind}$_{\text{CVPR}23}$ } &\textcolor{lightgray}{ 18.3 } & \textcolor{lightgray}{ 18.2} & \textcolor{lightgray}{ N/A} &\textcolor{lightgray}{  N/A} &\textcolor{lightgray}{  20.2}  & \textcolor{lightgray}{ 19.8}  & \textcolor{lightgray}{ N/A }& \textcolor{lightgray}{ N/A }   \\
\midrule
CAVMAE~\cite{gong2023contrastive}$_{\text{ICLR}23}$ & 21.6  & 20.7 & N/A & N/A  & 17.6  & 19.4 & N/A & N/A \\
DenseAV~\cite{hamilton2024separating}$_{\text{CVPR}24}$        
      & 30.3  & 23.0  & 30.9  & 23.3    & \textbf{43.0}  & 32.8 & \textbf{47.7} & \textbf{36.2} 
\\
\textbf{Ours}
   & \textbf{36.7}  & \textbf{27.0} & \textbf{32.8} & \textbf{24.3}   & 42.1 & \textbf{32.8} & 46.7 & 35.4         \\ 
\bottomrule
\end{tabular}
}
\caption{\textbf{Audio-Visual semantic segmentation with individual audios.}}
\vspace{-6mm}
\label{tab:clean_semseg_results}
\end{table}

\newpara{Results on speech prompted visual grounding.} Similar to the previous experiment, we evaluate our model using individual speech only, with results shown in~\Tref{tab:clean_semseg_results}. These results indicate that our method performs approximately similar to Hamilton~\etal~\cite{hamilton2024separating}, though slightly lower, on the speech-prompted segmentation task. This outcome aligns with our expectations because our approach enhances the model's understanding of non-speech sounds, allowing the sound head to specialize more effectively in non-speech audio without compromising the speech head’s ability to accurately interpret spoken language.

\begin{table}[t]
\centering
\resizebox{1.0\linewidth}{!}{  
\begin{tabular}{lcccccccc}
\toprule
\textbf{Method} & \multicolumn{8}{c}{\textbf{Clean Retrieval R@10}} \\ \cmidrule(lr){2-9}
& \multicolumn{4}{c}{\textbf{AudioSet}} & \multicolumn{4}{c}{\textbf{Places}} \\ \cmidrule(lr){2-5} \cmidrule(lr){6-9}
&\multicolumn{2}{c}{\textbf{Total}} &\multicolumn{2}{c}{\textbf{Sound}}
&\multicolumn{2}{c}{\textbf{Total}} &\multicolumn{2}{c}{\textbf{Speech}} \\
& \textbf{I$\rightarrow$A} & \textbf{A$\rightarrow$I} & \textbf{I$\rightarrow$A} & \textbf{A$\rightarrow$I} & \textbf{I$\rightarrow$A} & \textbf{A$\rightarrow$I} & \textbf{I$\rightarrow$A} & \textbf{A$\rightarrow$I} \\ 
\midrule
\textcolor{lightgray}{\textit{Foundation Model} }&  &  &  &  &  &  &  &    \\
\textcolor{lightgray}{ImageBind~\cite{girdhar2023imagebind}$_{\text{CVPR}23}$} & \textcolor{lightgray}{66.8}  & \textcolor{lightgray}{65.0}  & \textcolor{lightgray}{N/A}  & \textcolor{lightgray}{N/A}
& \textcolor{lightgray}{0.9}  & \textcolor{lightgray}{1.1} & \textcolor{lightgray}{N/A} & \textcolor{lightgray}{N/A}   \\
\midrule
CAVMAE~\cite{gong2023contrastive}$_{\text{ICLR}23}$
& 43.1  & 31.4 & N/A & N/A   
& 1.0  & 1.0 & N/A  & N/A \\
DenseAV~\cite{hamilton2024separating}$_{\text{CVPR}24}$     
& 41.8  & 43.6  & 46.7  & 46.5          
& \textbf{92.7}  & 92.4 & 93.8 & 93.9    
\\
\textbf{Ours}
& \textbf{45.5}  & \textbf{46.6} & \textbf{51.2} & \textbf{50.0}       
& 92.0 & \textbf{93.1} & \textbf{94.0} & \textbf{94.0}        
\\ 
\bottomrule
\end{tabular}
}
\caption{\textbf{Cross-modal retrieval task on Places and AudioSet.}}
\vspace{-6mm}
\label{tab:clean_retrieval_results}
\end{table}

\subsubsection{Cross-Modal Retrieval} \label{ssec:retrieval}
After demonstrating our method’s capabilities in both simultaneous and standard audio-visual grounding tasks, we also evaluate its performance on cross-modal retrieval.

\newpara{Implementation.} This experiment follows the same evaluation protocol of~\cite{hamilton2024separating}. We assess cross-modal retrieval using the PlacesAudio and AudioSet test sets as described in~\Sref{ssec:datasets}, using cross-modal retrieval accuracy@10 as the evaluation metric. For this experiment, the total aggregation approach involves summing across the heads, as in~\cite{hamilton2024separating}, different than individual audio-visual grounding experiments.

\newpara{Results.} Learning audio-visual associations is crucial for sound source localization. Therefore, models that perform well in sound source localization are also expected to do well in cross-modal retrieval~\cite{senocak2024aligning}, as they learn how objects look and sound, or how objects appear when referenced by spoken words.~\Tref{tab:clean_retrieval_results} shows the results for cross-modal retrieval tasks. As expected, our method significantly outperforms DenseAV in cross-modal retrieval on the AudioSet for both Audio→Image and Image→Audio tasks, achieving substantial improvements in both \textit{Total} and \textit{Sound Head} evaluations. On the Places dataset, our method performs almost identically to DenseAV across all experiments and head evaluations, following the same trend observed in the standard audio-visual segmentation task (~\Sref{ssec:grounding}). 
We note that our model is outperformed by ImageBind~\cite{girdhar2023imagebind} in cross-modal retrieval on AudioSet. This result is expected, given that ImageBind is a foundational model trained on large-scale datasets spanning multiple modalities, including text. However, ImageBind performs poorly on the Places dataset, whereas our method shows strong performance, underscoring its ability to effectively handle both types of audio.

\begin{table}[t]
\centering
\resizebox{1.0\linewidth}{!}{  
\begin{tabular}{lcccccccc}
\toprule
\textbf{Method} & \multicolumn{8}{c}{\textbf{Mixed SemSeg}} \\ \cmidrule(lr){2-9}
& \multicolumn{4}{c}{\textbf{Sound SemSeg}} & \multicolumn{4}{c}{\textbf{Speech SemSeg}} \\ \cmidrule(lr){2-5} \cmidrule(lr){6-9}
&\multicolumn{2}{c}{\textbf{Total}} &\multicolumn{2}{c}{\textbf{Sound}}
&\multicolumn{2}{c}{\textbf{Total}} &\multicolumn{2}{c}{\textbf{Speech}} \\
& \textbf{mAP} & \textbf{mIoU} & \textbf{mAP} & \textbf{mIoU} & \textbf{mAP} & \textbf{mIoU} & \textbf{mAP} & \textbf{mIoU} \\ 
\midrule      
\textcolor{lightgray}{\textit{Foundation Model}} &  &  &  &  &  &  &  &    \\
\textcolor{lightgray}{ImageBind~\cite{girdhar2023imagebind}$_{\text{CVPR}23}$  }& \textcolor{lightgray}{18.8 } & \textcolor{lightgray}{18.5} & \textcolor{lightgray}{N/A} & \textcolor{lightgray}{N/A} & \textcolor{lightgray}{19.3 } & \textcolor{lightgray}{19.5 } & \textcolor{lightgray}{N/A} & \textcolor{lightgray}{N/A  }  \\
\midrule
CAVMAE~\cite{gong2023contrastive}$_{\text{ICLR}23}$ & 19.6  & 18.9 & N/A & N/A  & 18.6  & 19.9 & N/A & N/A \\
DenseAV~\cite{hamilton2024separating}$_{\text{CVPR}24}$         
& 19.6  & 19.0  & 18.7  & 18.6    
& 30.5  & 24.2 & 33.0 & 25.6    
\\
\textbf{Ours}         
& \textbf{22.3}  & \textbf{20.2} & \textbf{28.4} & \textbf{22.9}      
& \textbf{39.3} & \textbf{29.5} & \textbf{41.6} & \textbf{31.2}  
 \\ 
\bottomrule
\end{tabular}
}
\caption{\textbf{Standard Audio-Visual semantic segmentation with mixture of audios.}}
\vspace{-4mm}
\label{tab:mixed_semseg_results}
\end{table}
\begin{table}[t]
\centering
\resizebox{1.0\linewidth}{!}{  
\begin{tabular}{lcccccccc}
\toprule
\textbf{Method} & \multicolumn{8}{c}{\textbf{Mixed Retrieval R@10}} \\ \cmidrule(lr){2-9}
& \multicolumn{4}{c}{\textbf{AudioSet}} & \multicolumn{4}{c}{\textbf{Places}} \\ \cmidrule(lr){2-5} \cmidrule(lr){6-9}
&\multicolumn{2}{c}{\textbf{Total}} &\multicolumn{2}{c}{\textbf{Sound}}
&\multicolumn{2}{c}{\textbf{Total}} &\multicolumn{2}{c}{\textbf{Speech}} \\
& \textbf{I$\rightarrow$A} & \textbf{A$\rightarrow$I} & \textbf{I$\rightarrow$A} & \textbf{A$\rightarrow$I} & \textbf{I$\rightarrow$A} & \textbf{A$\rightarrow$I} & \textbf{I$\rightarrow$A} & \textbf{A$\rightarrow$I} \\ 
\midrule      
\textcolor{lightgray}{\textit{Foundation Model}} &  &  &  &  &  &  &  &    \\
\textcolor{lightgray}{ImageBind~\cite{girdhar2023imagebind}$_{\text{CVPR}23}$ }
& \textcolor{lightgray}{35.5}  & \textcolor{lightgray}{35.0  }& \textcolor{lightgray}{N/A } & \textcolor{lightgray}{N/A }       
& \textcolor{lightgray}{1.20 } & \textcolor{lightgray}{0.70 }&\textcolor{lightgray}{ N/A }&\textcolor{lightgray}{ N/A }  \\
\midrule
CAVMAE~\cite{gong2023contrastive}$_{\text{ICLR}23}$
& \textbf{19.1}  & 8.9 & N/A & N/A   
& 0.8  & 1.3  & N/A  & N/A \\
DenseAV~\cite{hamilton2024separating}$_{\text{CVPR}24}$  
& 12.3  & 10.4  & 14.4  & 13.4           
& 49.6  & 54.2 & 50.6 & 58.7      
\\
\textbf{Ours}
& 17.8  & \textbf{20.3} & \textbf{44.3} & \textbf{42.6}     
& \textbf{79.0} & \textbf{82.8} & \textbf{87.3} & \textbf{86.2}        
\\ 
\bottomrule
\end{tabular}
}
\caption{\textbf{Cross-modal retrieval task on Places and AudioSet with mixture of audios.}}
\vspace{-6mm}
\label{tab:mixed_retrieval_results}
\end{table}

\subsubsection{Standard Experiments with Mixed Audio} \label{ssec:mixed}
Inspired by Mo \etal~\cite{mo2022SLAVC}, we added sound or spoken utterances from non-visible (\ie, off-screen) objects to the DenseAV datasets used in our experiments in~\Sref{ssec:grounding} and~\Sref{ssec:retrieval}. This addition creates a noisy dataset with mixed audio types. In this scenario, we can observe the robustness of the models to different audio types, as well as the degree of disentanglement—that is, how effectively each head is specialized only for its corresponding type of audio-visual association. Note that this setup differs from simultaneous localization experiments in two ways: (1) In the simultaneous setup, both audio types in the mixture have visual counterparts in the image, whereas here, the added audio of the opposite type is off-screen. (2) In simultaneous audio-visual segmentation, the model attempts to localize both audio types at once. In this setup, however, the model is tasked with localizing or retrieving information using only the target audio type, while ignoring any off-screen audio that lacks a visual counterpart.

\newpara{Implementation.} We repeat the experiments from~\Sref{ssec:grounding} and~\Sref{ssec:retrieval} — \textit{audio-visual segmentation} and \textit{cross-modal retrieval} - using identical implementation details, except that we employ a mixture of audio types as described above.

\newpara{Results.} Experimental results are presented in~\Tref{tab:mixed_semseg_results} and~\Tref{tab:mixed_retrieval_results}. As shown, our method significantly outperforms DenseAV across both tasks and evaluation approaches within each task. While the performance of both models declines with mixed audio inputs, the drop in performance is much smaller for our method (compare \Tref{tab:clean_semseg_results} \vs ~\Tref{tab:mixed_semseg_results} and~\Tref{tab:clean_retrieval_results} \vs ~\Tref{tab:mixed_retrieval_results}). We hypothesize that this performance gap and the smaller drop ratios with our method, compared to clean audio inputs, are due to two key factors: (1) Our model is more robust when processing mixed audio inputs due to the mix-and-separate approach and our joint learning objectives. (2) DenseAV's specialized heads may retain some information from the opposite audio type, which causes disturbances in mixed audio scenarios, while our heads are more effectively disentangled and specialized for their intended audio type. Also, CAV-MAE performs better in the \textit{Total Head} evaluation for I→A in AudioSet, but our specialized \textit{Sound Head} outperforms it significantly (44.3 \vs 19.1) and should be considered the target for this task, as Total Head is an alternative evaluation approach.

\subsubsection{Disentanglement Assessment} \label{ssec:disentanglement}
The success of the model using different heads for the target task is highly dependent on how well each head is specialized for its respective type, in other words, the level of disentanglement. Hamilton \etal~\cite{hamilton2024separating} evaluate disentanglement in two ways. First, they measure whether a head’s average activation strength can reliably indicate if a given sample predominantly contains ``language'' or ``sound'' - \textit{Pred.Dis}. Second, they assess the frequency with which the ``sound'' head incorrectly activates when only the ``language'' head should be active, and vice versa - \textit{Act.Dis}. We followed the evaluation protocols from~\cite{hamilton2024separating} exactly. The results for Pred.Dis. are 99.92\% for ours \vs 99.91\% for DenseAV, and for Act.Dis., 86.90\% for ours \vs 84.48\% for DenseAV. As seen, the performances are almost identical, as this approach merely acts as a selection mechanism for each type. However, depending on the level of disentanglement, each head may still retain some information from the opposite type of audio (\ie, information leakage), which could hinder the model's performance. To assess how much information each head still retains, and indirectly evaluate the disentanglement ability, we perform retrieval tasks using the opposite audio-type heads.
\begin{table}[t]
\centering
\resizebox{1.0\linewidth}{!}{  
\begin{tabular}{lcccccccc}
\toprule
\multirow{11.4}{*}{\textbf{Method}} & \multicolumn{8}{c}{\textbf{Retrieval R@10}} \\ \cmidrule(lr){2-9}
& \multicolumn{4}{c}{\textbf{Clean}} & \multicolumn{4}{c}{\textbf{Mixed}} \\ \cmidrule(lr){2-5} \cmidrule(lr){6-9}
& \multicolumn{2}{c}{\textbf{Places}} & \multicolumn{2}{c}{\textbf{AudioSet}} & \multicolumn{2}{c}{\textbf{Places}} & \multicolumn{2}{c}{\textbf{AudioSet}} \\ \cmidrule(lr){2-3} \cmidrule(lr){4-5} \cmidrule(lr){6-7} \cmidrule(lr){8-9}
&\multicolumn{2}{c}{\textbf{Sound}} &\multicolumn{2}{c}{\textbf{Speech}}
&\multicolumn{2}{c}{\textbf{Sound}} &\multicolumn{2}{c}{\textbf{Speech}} \\
& \textbf{I$\rightarrow$A} & \textbf{A$\rightarrow$I} & \textbf{I$\rightarrow$A} & \textbf{A$\rightarrow$I} & \textbf{I$\rightarrow$A} & \textbf{A$\rightarrow$I} & \textbf{I$\rightarrow$A} & \textbf{A$\rightarrow$I} \\ 

\midrule
DenseAV& 13.5 & 31.6 & \textbf{3.4} & \textbf{6.9} & 4.4 & 12.6 & 1.8 & 1.5 \\
\textbf{Ours} & \textbf{2.3} & \textbf{10.5} & 3.7 & 8.0 & \textbf{1.4} & \textbf{1.4} & \textbf{1.6} & \textbf{1.3} \\ 
\bottomrule
\end{tabular}
}
\caption{\textbf{Measuring disentanglement.} Each head is tasked with retrieval by using the opposite audio type to reveal retained information within each head. A lower number is better performance.}
\vspace{-6mm}
\label{tab:disentangle_assesment}
\end{table}

\newpara{Results.} As shown in~\Tref{tab:disentangle_assesment}, our method generally yields lower performance compared to DenseAV, which suggests that the heads produced by our method contain less information about the opposite type of audio. Additionally, we observe that our proposed approach has improved the specialization of the sound head more than the speech head, likely because the audio backbone network is already optimized for speech. Overall, these results demonstrate the effectiveness of our approach for improving disentanglement.

\subsection{Ablation Results} \label{ssec:ablation}
\begin{table}[t!]
\footnotesize
\renewcommand{\arraystretch}{1.3}
\centering
    \resizebox{0.45\linewidth}{!}{
    \begin{tabular}{lcc}
    \hline
    {} & \multicolumn{2}{c}{\textbf{AudioSet}} \\
    \cmidrule{2-3}
    \textbf{Method} & \textbf{I$\rightarrow$A} & \textbf{A$\rightarrow$I} \\
    \hline
    $L_{dis}$  & 30.9 & 32.0 \\
    \textbf{Ours} & \textbf{51.2} & \textbf{50.0}  \\
    \hline
    \multicolumn{3}{c}{(a)} \\ 
    \end{tabular}}
    \hspace{5pt}
    \resizebox{0.45\linewidth}{!}{
    \begin{tabular}{lcc}
    \hline
    {} & \multicolumn{2}{c}{\textbf{Places}} \\
    \cmidrule{2-3}
    \textbf{Method} & \textbf{I$\rightarrow$A} & \textbf{A$\rightarrow$I} \\
    \hline
    $L_{cor}$  &  23.7 & 42.1 \\
    \textbf{Ours} & \textbf{2.3} & \textbf{10.5} \\
    \hline
    \multicolumn{3}{c}{(b)} \\ 
    \end{tabular}}\\
    \vspace{-2mm}
\caption{\textbf{Ablation results for training losses.}}
\vspace{-4mm}
\label{tab:ablation}
\end{table}

We conduct experiments to verify that each of our audio-visual alignment objectives — correspondence and disentanglement — functions as intended. Using the ``Sound Head'' of our model for retrieval tasks throughout this ablation study, we first assess the effect of omitting the correspondence loss by measuring retrieval performance on AudioSet, which targets the same audio type. As shown in~\Tref{tab:ablation}(a), omitting the correspondence loss leads to a \app20\% drop in performance, confirming that this loss effectively enables audio-visual matching necessary for cross-modal retrieval. In the second experiment, we examine the impact of omitting the disentanglement loss by evaluating ``Sound Head'' performance on the Places dataset, which targets the opposite audio type. Here, we expect poor disentanglement to cause the Sound Head to resemble the Speech Head, yielding high retrieval performance for speech. Results in~\Tref{tab:ablation}(b) show a \app22\% performance increase for the ``Sound Head'', indicating it retains speech information due to inadequate disentanglement. In summary, these ablation studies confirm that our learning objectives perform as designed.

\vspace{-2mm}
\section{Conclusion}
In this work, we introduce the first unified model for simultaneous grounding of mixed audio types — including spoken language and non-speech sounds — within a visual scene. Using a novel ``mix-and-separate'' framework with joint correspondence and disentanglement objectives, our approach learns distinct representations for each audio type, addressing key limitations of prior models limited to handling single or sequential audio sources. This framework enables our model to produce accurate, distinct embeddings for each audio type, achieving robust disentanglement and grounding even with overlapping mixed audio. We also present a new dataset for evaluating simultaneous audio-visual grounding with mixed audio inputs, where our model outperforms previous methods across this and other segmentation and retrieval tasks. In summary, our contributions establish a new paradigm for audio-visual grounding in multi-source environments, advancing machine perception capabilities for real-world auditory-visual interactions.
\vspace{-2mm}
\section{Acknowledgment}
This work was supported by the National Research Foundation of Korea (NRF) grant funded by the Korea government (MSIT) (No. RS-2023-00212845, Multimodal Speech Processing for Human-Computer Interaction).
{
    \small
    \bibliographystyle{ieeenat_fullname}
    \bibliography{main}
}

\clearpage
\setcounter{page}{1}
\maketitlesupplementary

The contents in this supplementary material are as follows:

Details on 
Extended IS3 Dataset (\Sref{sec:a}), 
Re-visiting Heads (\Sref{sec:b}), 
Additional Ablation Results (\Sref{sec:c}), 
Computational Overhead (\Sref{sec:d}),
Additional Qualitative Results (\Sref{sec:e}), 
Performance Discrepancy with DenseAV (\Sref{sec:f})
and Implementation Details (\Sref{sec:g}).

\section{Details on Extended IS3 Dataset}\label{sec:a}
We extend the IS3 dataset~\cite{senocak2024aligning} to enable simultaneous grounding of mixture of audios.
The dataset is originally designed for interactive sound source localization, consisting of 3,420 images, each paired with two general sound samples corresponding to two visible objects in the scene. We generate speech samples for each visible object in each image. These speech samples, created using the Google gTTS API, read the class categories of the two visible objects. 
Then each image contains two objects, each associated with one sound and one speech sample. We form triplets by combining the image with the sound from one object and the speech from the other, and vice versa. The sound and speech samples of a triplet are then mixed together to form a single combined auditory input. This extension enables simultaneous grounding of mixed audio types which requires disentangling overlapping auditory inputs—sound and speech—and aligning them with the correct visual objects. The Extended IS3 dataset thus serves as a comprehensive benchmark for evaluating the capability on audio-visual interactions in real-world scenarios.

\section{Re-visiting Heads}\label{sec:b}
As outlined in our architecture, the audio encoder includes two specialized heads: one for speech and one for sound. Additionally, in~\Sref{ssec:grounding} of the main paper, we discuss different evaluation approaches for these heads, namely \textit{Specialized Heads} and \textit{Total Head}. Here, we would like to re-emphasize that the primary focus of the evaluation should be on the specialized heads, depending on the target task. The \textit{Total Head}, introduced by Hamilton \etal~\cite{hamilton2024separating}, serves as an alternative evaluation approach. In summary, tasks such as segmentation and retrieval on the Places dataset should use the \textit{Speech Head}, while all tasks on the AudioSet dataset should utilize the \textit{Sound Head}.

As shown in~\Tref{tab:clean_retrieval_results} and~\Tref{tab:mixed_retrieval_results} of the main paper, the \textit{Total Head} approach exhibits performance degradation on benchmarks such as AudioSet and Places. This performance gap is especially evident in~\Tref{tab:mixed_retrieval_results}, where the mixed audio scenario highlights the limitations of aggregating unrelated similarity volumes (\textit{Total Head} approach). The inclusion of irrelevant information introduces noise, leading to considerably worse results compared to the \textit{Specialized Head} approach, which focuses solely on the relevant audio type and achieves better performance. Similarly, as mentioned in the main paper (\Sref{ssec:mixed}), CAV-MAE performs better than ours in the \textit{Total Head} for I$\rightarrow$A on AudioSet. However, the target for this task should be the \textit{Sound Head}, and the performance degradation from \textit{Sound Head} to \textit{Total Head} can serve as evidence of disentanglement ability of our model, as the \textit{Total Head} introduces noise from unrelated sub-modalities.

To highlight the scenario where the \textit{Total Head} is particularly useful, we conducted an additional novel task on the Extended IS3 dataset.
This task involves performing retrieval using audio inputs that combine speech and general sounds, along with images containing two related objects. In this case, using the \textit{Total Head} is more appropriate as it better reflects the characteristic of the dataset and the task.
We present results in~\Tref{tab:simul_ablate} by comparing head selection methods. \textit{Total} uses summation across sub-modalities:
\begin{align}
S_{sum}(\mathbf{a}, \mathbf{v}) &= \sum_{k} \left( S(\mathbf{a}, \mathbf{v}) \right),
\label{eq:selection_methods_summation}
\end{align}
where $k$ refers the number of audio 
types. \textit{Sound Head} and \textit{Speech Head} directly apply the head selection approach in~\eqref{eq:selection_methods_indexing}. \textit{Total Head} evaluation outperforms both specialized heads on the Extended IS3 dataset by approximately 10\% or more. 

\begin{table}[h]
\centering
\resizebox{1.0\linewidth}{!}{  
\begin{tabular}{lcccccc}
\toprule
& \multicolumn{6}{c}{\textbf{Retrieval R@10}} \\ \cmidrule(lr){2-7}
& \multicolumn{2}{c}{\textbf{Total}} &\multicolumn{2}{c}{\textbf{Sound}} 
& \multicolumn{2}{c}{\textbf{Speech}} \\
\textbf{Method} & \textbf{I$\rightarrow$A} & \textbf{A$\rightarrow$I}  & \textbf{I$\rightarrow$A} & \textbf{A$\rightarrow$I} & \textbf{I$\rightarrow$A} & \textbf{A$\rightarrow$I} \\ 
\midrule
DenseAV~\cite{hamilton2024separating}$_{\text{CVPR}24}$
 & 25.2 & 11.7 &  14.7 & 9.7 & 19.0 & 9.9\\
\textbf{Ours} 
& \textbf{29.7} & \textbf{25.0} &  \textbf{18.3} & \textbf{11.9} & \textbf{19.8} & \textbf{16.8} \\ 
\bottomrule
\end{tabular}
}
\caption{\textbf{Cross-modal retrieval task on Extended IS3.}}
\vspace{-6mm}
\label{tab:simul_ablate}
\end{table}

\section{Additional Ablation Results}\label{sec:c}
In~\Sref{ssec:ablation} of the main paper, we presented ablation results to evaluate the impact of our audio-visual alignment objectives—correspondence and disentanglement. Here, we provide additional results on the retrieval task for both clean and mixed audio cases to present a more comprehensive analysis. Results are in~\Tref{tab:clean_retrieval_results_supp} and~\Tref{tab:mixed_retrieval_results_supp}. 

Firstly, as shown in~\Tref{tab:clean_retrieval_results_supp}, omitting the disentanglement loss causes the \textit{Speech Head} to perform numerically higher (indicating worse performance) on AudioSet by approximately 30\%, and the \textit{Sound Head} to perform worse on Places by over 20\% in clean retrieval. This suggests that the opposite heads are being activated by the incorrect audio type, which is undesirable. A similar trend is observed in the mixed retrieval results in~\Tref{tab:mixed_retrieval_results_supp}, although the performance gap is slightly smaller. These findings indicate that without the disentanglement loss, the heads fail to specialize effectively for their intended roles.
Secondly, we examine the impact of omitting the correspondence loss by evaluating the specialized head for each dataset. In both~\Tref{tab:clean_retrieval_results_supp} and~\Tref{tab:mixed_retrieval_results_supp}, 
performance drops by over 15\% on AudioSet and \app 3\% on Places, confirming that $L_{cor}$ effectively enables audio-visual matching necessary for cross-modal retrieval. 
It is noteworthy that the model trained only with $L_{cor}$ struggles with mixed audio, as $L_{dis}$ provides robustness against noise from the opposite audio type due to efficient disentanglement.

\begin{table}[h]
\centering
\resizebox{1.0\linewidth}{!}{  
\begin{tabular}{lcccccccccccc}
\toprule
& \multicolumn{12}{c}{\textbf{Clean Retrieval R@10}} \\ \cmidrule(lr){2-13}&
\multicolumn{6}{c}{\textbf{AudioSet}} & \multicolumn{6}{c}{\textbf{Places}} \\ \cmidrule(lr){2-7} \cmidrule(lr){8-13}
&\multicolumn{2}{c}{\textbf{Total $\uparrow$}} &\multicolumn{2}{c}{\textbf{Sound $\uparrow$}}
&\multicolumn{2}{c}{\textbf{Speech $\downarrow$}}
&\multicolumn{2}{c}{\textbf{Total $\uparrow$}} &\multicolumn{2}{c}{\textbf{Sound $\downarrow$}}
&\multicolumn{2}{c}{\textbf{Speech $\uparrow$}} \\
\textbf{Method}& \textbf{I$\rightarrow$A} & \textbf{A$\rightarrow$I} & \textbf{I$\rightarrow$A} & \textbf{A$\rightarrow$I} & \textbf{I$\rightarrow$A} & \textbf{A$\rightarrow$I} & \textbf{I$\rightarrow$A} & \textbf{A$\rightarrow$I} & \textbf{I$\rightarrow$A} & \textbf{A$\rightarrow$I} & \textbf{I$\rightarrow$A} & \textbf{A$\rightarrow$I} \\ 
\midrule
$L_{cor}$
& \textbf{48.6} & \textbf{47.4}  & 34.0 & 35.8 & 38.3 & 38.7 & \textbf{92.3} & 91.9 & 23.7 & 42.1 & 92.0 & 92.5\\
$L_{dis}$
& 29.2 & 28.5 & 30.9 & 32.0 & 4.7  & \textbf{7.1}  &  91.5 & 90.5 & \textbf{2.1} & \textbf{6.3} & 91.1 & 91.2 \\
\textbf{Ours}
& {45.5}  & {46.6} & \textbf{51.2}  & \textbf{50.0} & \textbf{3.7}  & 8 & 92.0 & \textbf{93.1} & 2.3 & 10.5 
& \textbf{94.0} & \textbf{94.0}        
\\ 
\bottomrule

\end{tabular}
}
\caption{\textbf{Cross-modal retrieval task on Places and AudioSet.}}
\vspace{-4mm}
\label{tab:clean_retrieval_results_supp}
\end{table}

\begin{table}[h]
\centering
\resizebox{1.0\linewidth}{!}{  
\begin{tabular}{lcccccccccccc}
\toprule
& \multicolumn{12}{c}{\textbf{Mixed Retrieval R@10}} \\ \cmidrule(lr){2-13}&
\multicolumn{6}{c}{\textbf{AudioSet}} & \multicolumn{6}{c}{\textbf{Places}} \\ \cmidrule(lr){2-7} \cmidrule(lr){8-13}
&\multicolumn{2}{c}{\textbf{Total $\uparrow$}} &\multicolumn{2}{c}{\textbf{Sound $\uparrow$}}
&\multicolumn{2}{c}{\textbf{Speech $\downarrow$}}
&\multicolumn{2}{c}{\textbf{Total $\uparrow$}} &\multicolumn{2}{c}{\textbf{Sound $\downarrow$}}
&\multicolumn{2}{c}{\textbf{Speech $\uparrow$}} \\
\textbf{Method} & \textbf{I$\rightarrow$A} & \textbf{A$\rightarrow$I} & \textbf{I$\rightarrow$A} & \textbf{A$\rightarrow$I} & \textbf{I$\rightarrow$A} & \textbf{A$\rightarrow$I} & \textbf{I$\rightarrow$A} & \textbf{A$\rightarrow$I} & \textbf{I$\rightarrow$A} & \textbf{A$\rightarrow$I} & \textbf{I$\rightarrow$A} & \textbf{A$\rightarrow$I} \\ 
\midrule
$L_{cor}$
& 14.0 & 11.4  & 11.3 & 9.6 & 7.5 & 9.4 & 49.8 & 51.2 & 7.8 & 21.4 & 52.1 & 53.3 \\
$L_{dis}$
& \textbf{19.4} & 19.9 & 29.3 & 28.3 & 3.6  & 3.1  & 76.5 & 80.8 & 1.7 & 3.2 & 82.7 & 83.6 \\
\textbf{Ours}
& {17.8}  & \textbf{20.3} & \textbf{44.3}  & \textbf{42.6} & \textbf{1.6} & \textbf{1.3} & \textbf{79.0}  & \textbf{82.8} & \textbf{1.4} & \textbf{1.4} & \textbf{87.3} & \textbf{86.2}
\\ 
\bottomrule
\end{tabular}
}
\caption{\textbf{Cross-modal retrieval task on Places and AudioSet with mixture of audios.}}

\label{tab:mixed_retrieval_results_supp}
\vspace{-4mm}
\end{table}

\section{Computational Overhead}\label{sec:d}
\vspace{-2mm}
\begin{table}[H]
\footnotesize
\centering
\resizebox{0.5\linewidth}{!}{  
    \begin{tabular}{lcc}
    \hline
    \textbf{Model} & {DenseAV} & {Ours} \\
    \hline
    \textbf{FLOPs(G)} & 4151.25 & 12002.54 \\
    \hline
    \end{tabular}
}
\caption{\textbf{Computational overhead during training.}}
\vspace{-4mm}
\label{tab:suppl_computational_overhead}
\end{table}
Our approach uses multiple forward passes through the image and audio encoders to process clean audios, their paired images, and mixed
audio inputs (Figure~\ref{fig:framework_figure}), 
 increasing overhead during training compared to DenseAV~\cite{hamilton2024separating} (Table~\ref{tab:suppl_computational_overhead}). However, this overhead is only present during training, as the test phase is a single forward pass.

\section{ Additional Qualitative Results}
\label{sec:e}

\subsection{Segmentation Benchmarks}
\begin{figure*}
    \centering
    \captionsetup{type=figure}
    \includegraphics[width=\textwidth]{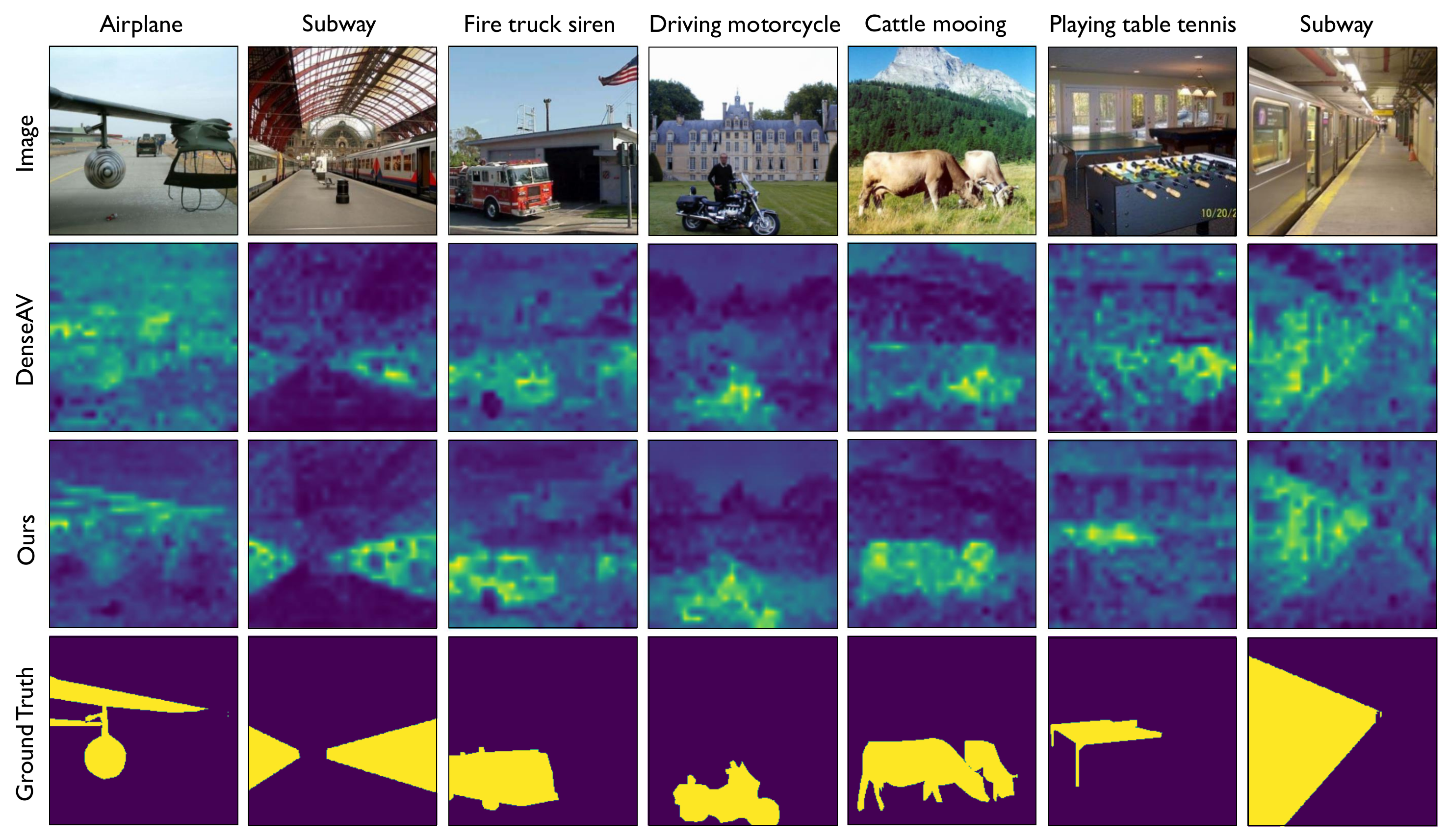}
    \caption{\textbf{Sound prompted semantic segmentation on dataset from~\cite{hamilton2024separating}.} 
    }
    \label{fig:qual_standard_sound}
\end{figure*}

\begin{figure*}
    \centering
    \captionsetup{type=figure}
    \includegraphics[width=0.99\textwidth]{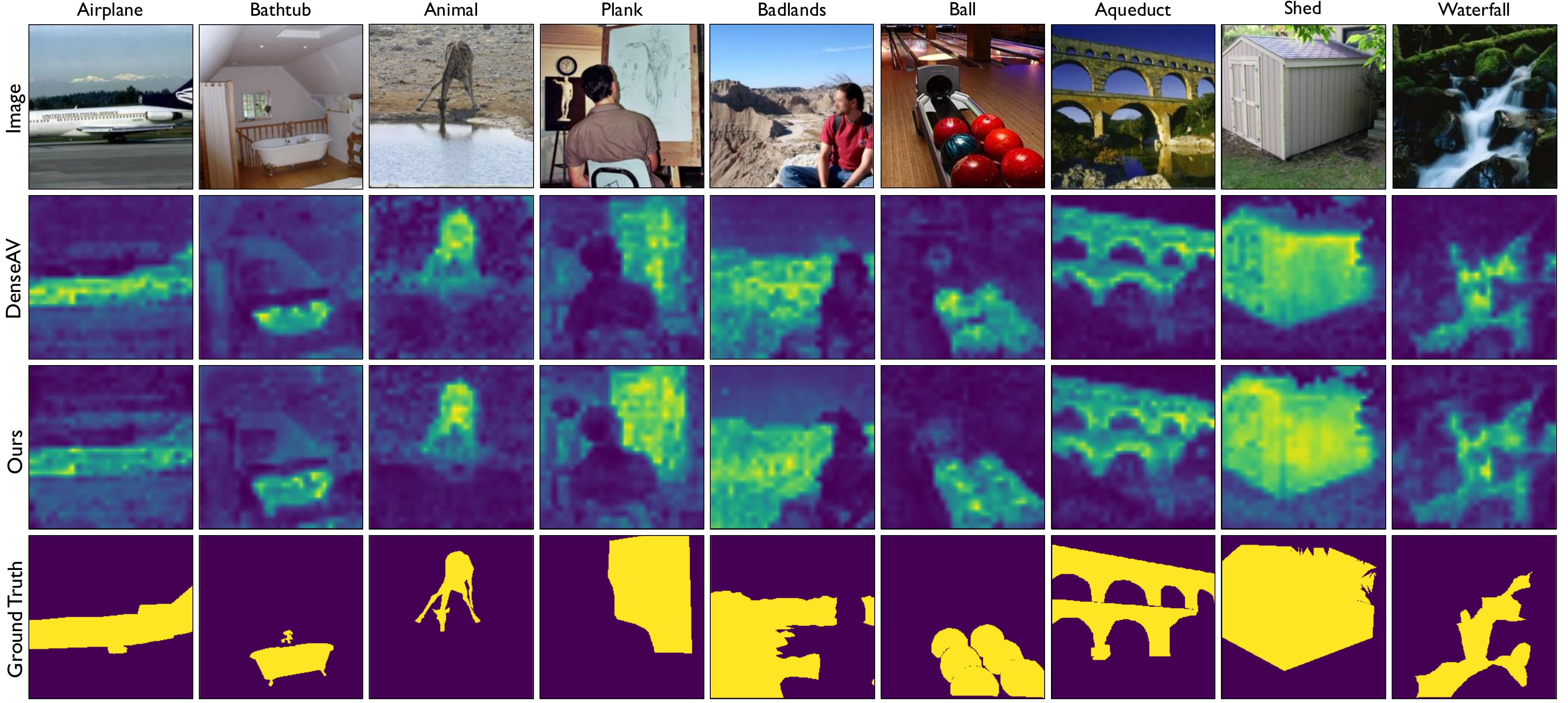}
    \caption{\textbf{Speech prompted semantic segmentation on dataset from~\cite{hamilton2024separating}.} 
    }
    \label{fig:qual_standard_speech}
\end{figure*}

\begin{figure*}[t]
    \centering
    \captionsetup{type=figure}
    \includegraphics[width=\textwidth]{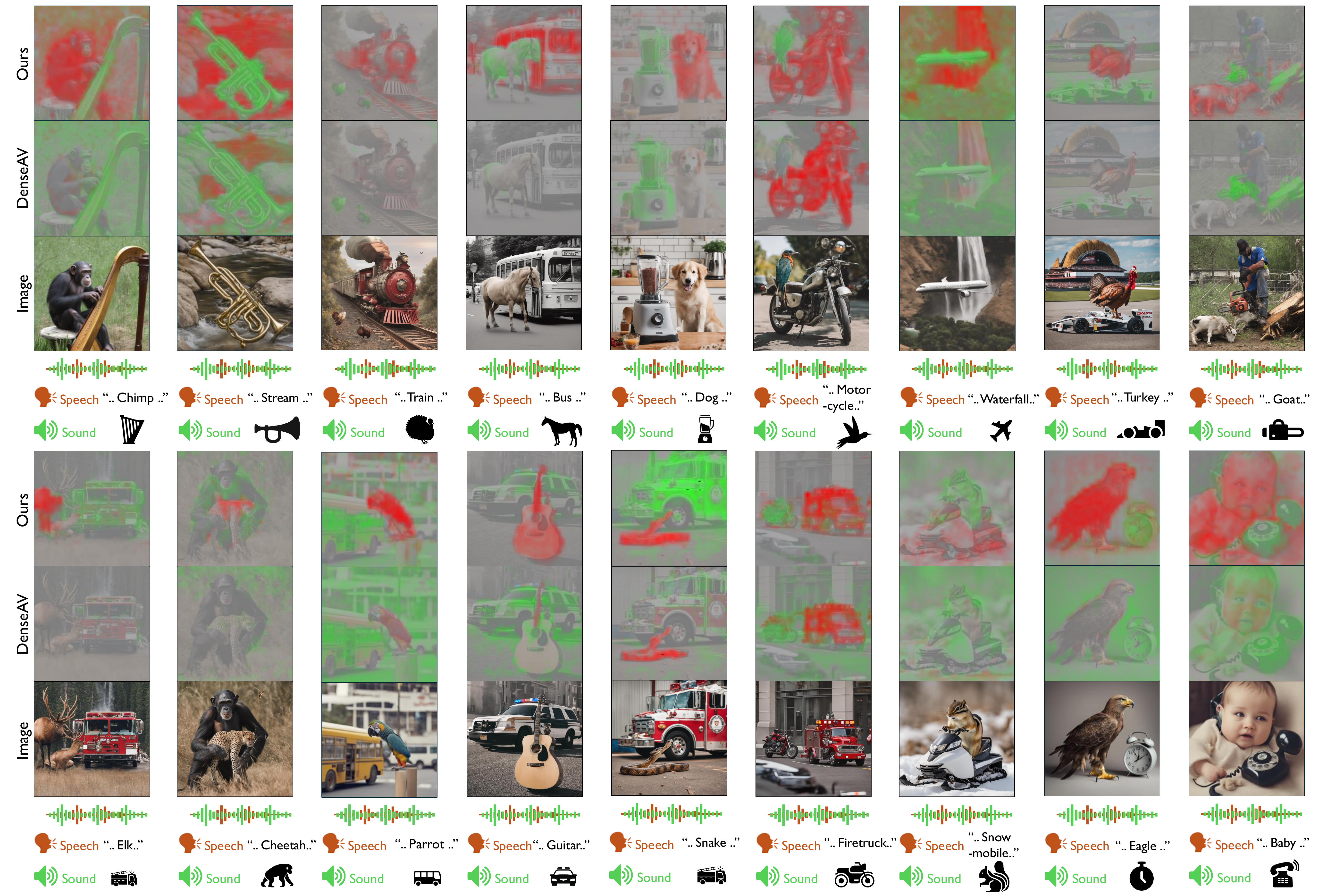}
    \caption{\textbf{Simultaneous semantic segmentation on Extended IS3} 
    }
    \label{fig:qual_simultaneous_semseg}
\end{figure*}

Due to space constraints in the main paper, we included only a limited number of qualitative results from the Simultaneous Segmentation experiment. In this supplementary material, we provide additional qualitative results for the standard segmentation task, including Sound-Prompted Semantic Segmentation, Speech-Prompted Semantic Segmentation, and Simultaneous Segmentation on the Extended IS3 dataset, as shown in~\Fref{fig:qual_standard_sound},~\Fref{fig:qual_standard_speech}, and~\Fref{fig:qual_simultaneous_semseg}, respectively.

\subsection{Real-world Scenarios}
\begin{figure*}
    \centering
    \captionsetup{type=figure}
    \includegraphics[width=0.90\textwidth]{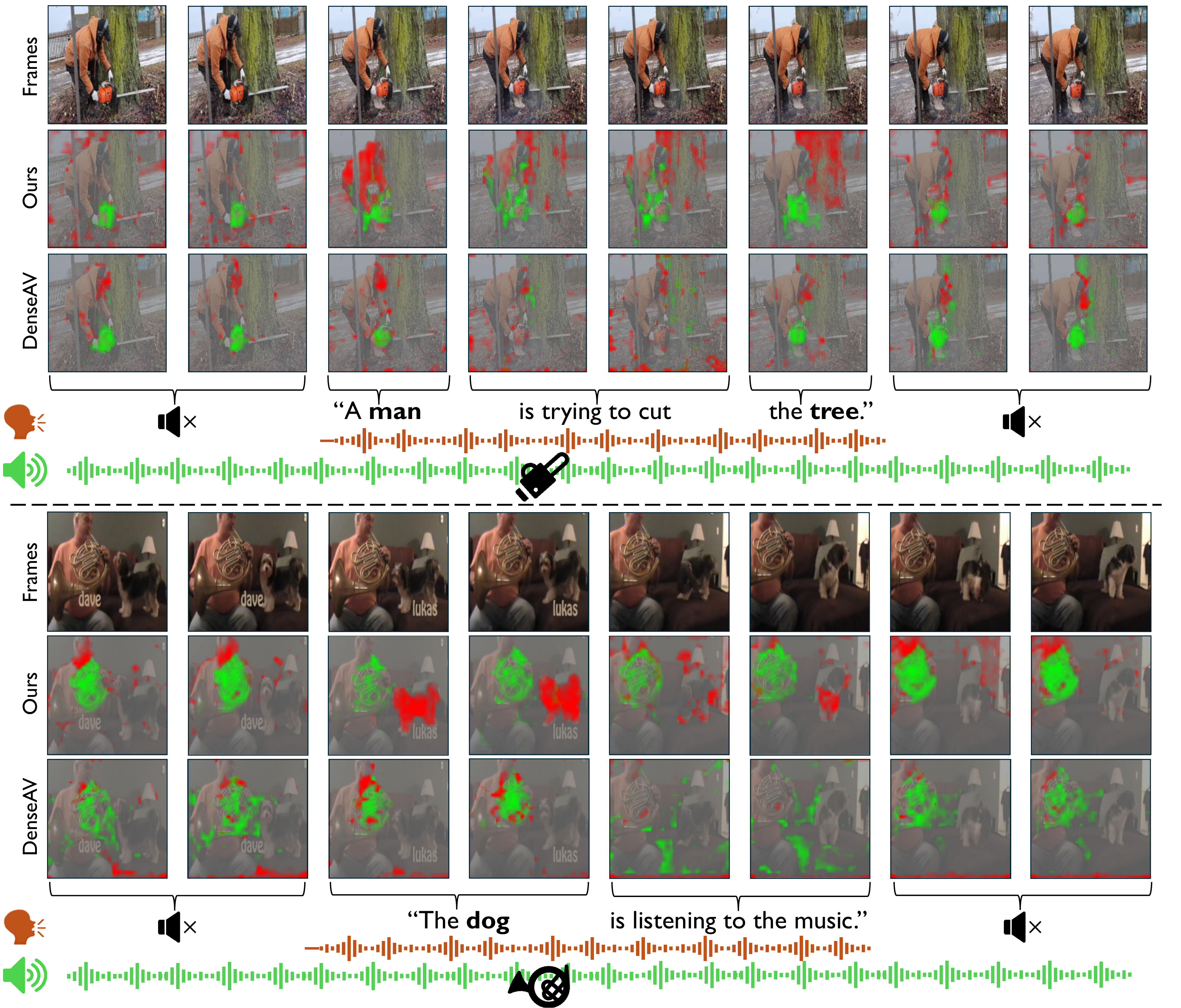}
    \caption{\textbf{Real-world scenarios.} 
    }
    \label{fig:qual_real_world_scenarios}
\end{figure*}

The examples in Figure \ref{fig:qual_real_world_scenarios} qualitatively compare our model with DenseAV in real-world scenarios from YouTube. When sound and speech overlap, our model grounds both the sound source and the object mentioned in speech more robustly. Our model demonstrates strong performance not only on evaluation dataset composed of TTS-generated speech but also on real-world speech samples.
\subsection{Failure Cases}

\begin{figure*}
    \centering
    \includegraphics[width=0.45\linewidth]{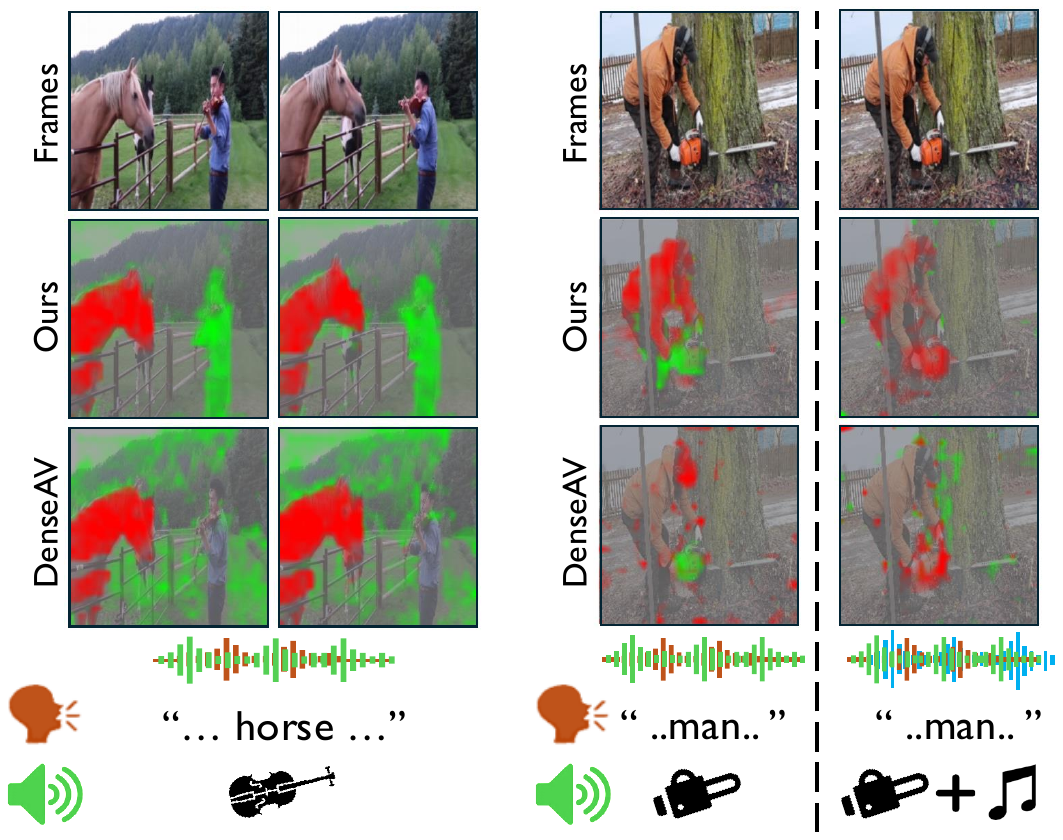}
    \caption{\textbf{Failure cases.}}
    \label{fig:failure_cases}
    \vspace{-2mm}
\end{figure*}

We present two failure cases: (1) When the interacted object is too small or occluded, the model may capture both the object and the person (Figure \ref{fig:failure_cases}, left). (2) When additional background noise or music increases the complexity of the audio, it makes localization harder (Figure \ref{fig:failure_cases}, right). Despite these challenges, our model remains more robust than DenseAV.

\section{Performance Discrepancy with DenseAV}\label{sec:f}
Our approach builds on DenseAV~\cite{hamilton2024separating} by introducing joint learning objectives (Mix-and-Separate approach), while strictly following the official DenseAV GitHub implementation without any modifications. All results related to DenseAV reported in our paper were obtained using the official DenseAV checkpoint.
The differences between our reported results and those presented in the original DenseAV paper can be attributed to two main factors.
(1) DenseAV proposes three settings: Sound-only, Speech-only, and Sound-and-Speech. Since the DenseAV paper suggests that a single model can distinguish both sound and speech, one might expect that all reported results use the Sound-and-Speech setting. However, most of the results in the original paper are based on the Sound-only and Speech-only settings, with the exception of the disentanglement evaluation. In contrast, our work focuses on simultaneous grounding of sound and speech within a single model, and thus we adopt the Sound-and-Speech setting as our baseline.
(2) The evaluation sample list for AudioSet used in the cross-modal retrieval task was not publicly available. To ensure fairness, we evaluated multiple random combinations of 1,000 AudioSet test samples. While the exact results could not be reproduced, we observed that performance was generally consistent across different splits. We therefore report results from one representative combination.
As a result of these differences in evaluation setup, the DenseAV scores reported in our segmentation and retrieval experiments may appear lower than those presented in the original paper. We include this clarification to help avoid potential confusion and to ensure fair and transparent comparison.

\section{Implementation Details}\label{sec:g}

\subsection{Regularizers}

We incorporate several regularization terms proposed by Hamilton et al.~\cite{hamilton2024separating} to improve training stability. We re-emphasize that while these techniques do not significantly impact the model's performance, they contribute to more stable training. 
\textbf{Disentanglement Regularizer} encourages different similarity volumes to specialize in distinct audio-visual associations by penalizing simultaneous activations across heads:
\begin{equation}
\label{eq:regularizer_disreg}
\mathcal{L}_{DisReg} = \text{Mean}(|S(\mathbf{a}_b, \mathbf{v}_b)[1] \circ S(\mathbf{a}_b, \mathbf{v}_b)[2]|),
\end{equation}
where $\circ$ denotes element-wise multiplication and $\mathbf{a}_b$, and $\mathbf{v}_b$ refers to audio and image feature of $b^{th}$ sample from a batch respectively. The \textbf{Stability Regularizer} consists of several smaller regularization terms. 
The \textbf{Negative Audio Splicing Regularizer} prevents self-attention mechanisms from collapsing by relying exclusively on specific tokens. It introduces negative audio regions into audio clips and penalizes activations in these regions. This is defined as:
\begin{equation}
\label{eq:regularizer_splice}
\mathcal{L}_{Splice} = \text{WeightedMean}(S(\mathbf{a}_b, \mathbf{v}_b)^2, m_b),
\end{equation}
where $m_b$ represents a soft mask identifying spliced regions. 
The \textbf{Calibration Regularizer} ensures that the calibration temperature $\tau$ remains stable by penalizing values over 1.0, expressed as:
\begin{equation}
\label{eq:regularizer_cal}
\mathcal{L}_{Cal} = \max(\log(\tau), 0)^2.
\end{equation}
The \textbf{Non-Negative Pressure Regularizer} promotes positive feature similarity by penalizing similarity scores below zero:
\begin{equation}
\label{eq:regularizer_nonneg}
\mathcal{L}_{NonNeg} = \frac{1}{|\Omega|} \sum_{\Omega} \min(S(\mathbf{a}_b, \mathbf{v}_{b'})[k,f,t,h,w], 0)^2,
\end{equation}
where $\Omega$ is a set of randomly selected coordinates from the similarity volumes and $b'$ refers to another sample from the batch. Lastly, the \textbf{Total Variation Smoothness Regularizer} ensures temporal consistency by penalizing rapid changes in activations over time, defined as:
\begin{equation}
\label{eq:regularizer_tv}
\mathcal{L}_{TV} = \text{Mean}((\text{act}(1:t-1) - \text{act}(2:t))^2),
\end{equation}
where activations over time are defined as $\text{act}(1:t-1) = (S(\mathbf{a}_b, \mathbf{v}_b)[:,:,t',:,:])_{t'=1}^{t-1}$. Combining these terms, Hamilton et al.~\cite{hamilton2024separating} defined the overall stability regularizer as:
\begin{equation}
\label{eq:regularizer_stability}
\begin{aligned}
\mathcal{L}_{Stability} = &\lambda_{Splice} \mathcal{L}_{Splice} + \lambda_{Cal}\mathcal{L}_{Cal} \\ + &\lambda_{NonNeg} \mathcal{L}_{NonNeg} + \lambda_{TV} \mathcal{L}_{TV},
\end{aligned}
\end{equation}
where $\lambda_{Splice} = 0.01$, $\lambda_{Cal} = 0.1$, $\lambda_{NonNeg} = 0.01$, and $\lambda_{TV} = 0.01$.

\end{document}